\documentclass{article}
\usepackage{pdflscape}
\usepackage{arxiv}
\usepackage{amsmath}
\usepackage[utf8]{inputenc} % allow utf-8 input
\usepackage[T1]{fontenc}    % use 8-bit T1 fonts
\usepackage[hidelinks]{hyperref}      % hyperlinks
\usepackage{url}            % simple URL typesetting
\usepackage{booktabs}       % professional-quality tables
\usepackage{amsfonts}       % blackboard math symbols
\usepackage{nicefrac}       % compact symbols for 1/2, etc.
\usepackage{microtype}      % microtypography
\usepackage{cleveref}  
\usepackage[ruled,vlined]{algorithm2e}
\usepackage{lipsum}         % Can be removed after putting your text content
\usepackage{tabularx}
\usepackage{longtable}
\usepackage{booktabs}
\usepackage{pdflscape}
\usepackage{makecell}
\usepackage{array}
\usepackage{graphicx}
\usepackage{ragged2e}
\newcolumntype{L}[1]{>{\RaggedRight\arraybackslash}p{#1}}
\usepackage[numbers]{natbib}
\usepackage{doi}

\title{A Transferable Learned Temporal Prior for Transmission Reconstruction and Decision-Relevant Uncertainty in Real Outbreak Labels}
\date{}

\newif\ifuniqueAffiliation

\uniqueAffiliationtrue

\ifuniqueAffiliation % Standard variant of author block
\author{ \href{https://orcid.org/0009-0001-5024-8448}{\includegraphics[scale=0.06]{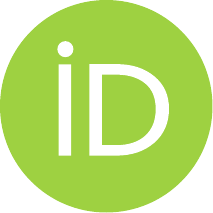}\hspace{1mm}Md Ahsan Karim} \\
	Department of Computer Science and Engineering, National Institute of Textile Engineering and Research (NITER)\\
	Nayarhat, Savar, Dhaka-1350, Bangladesh \\
	\texttt{makarim11@niter.edu.bd} \\
}

\renewcommand{\headeright}{
}
\renewcommand{\shorttitle}{Learned Temporal Priors for Outbreak Reconstruction}

\begin{document}
\maketitle

\begin{abstract}
Reconstructing who infected whom often relies on case timing and published
transmission links. However, two important questions remain: can a temporal
pattern learned from other diseases transfer to a new outbreak, and how
reliable are the transmission links used as ground truth? We learned a temporal prior from eleven disease groups using logistic
regression. The model was locked before any target-outbreak data were accessed
and was then tested without refitting on 29 Andes virus (ANDV) parent-ranking
tasks. The locked prior achieved a mean reciprocal rank (MRR) of 0.571,
compared with 0.274 for the strongest fair source-trained temporal baseline
(permutation $p<0.001$). Its Top-1 accuracy was 37.9\%, compared with 13.8\%.
The MRR advantage remained significant unless 7--8 favorable task outcomes
were reversed. We separately examined the reliability of published transmission links.
Among 75 epidemiologically linked inter-host pairs from the 2022 New York City
mpox outbreak, 54.67\% were genomically unresolved or unsupported as direct
transmission links (exact 95\% CI: 42.75--66.21\%). We also tested whether
retaining uncertain transmission edges changed source prioritization. In ANDV
and Guangdong Delta outbreak graphs, the top-5 priority sets changed, with
Jaccard similarity ranging from 0.429 to 0.667. These results show that temporal information learned across diseases can
provide a useful ranking signal without target-specific refitting. They also
show that uncertainty in published transmission links can affect which source
cases are prioritized for investigation.
\end{abstract}

% keywords can be removed

\section{Introduction}
\label{sec:1_Intro}

Tracking who infected whom is a critical task during infectious disease outbreaks. When health officials confirm a new case, they often need to identify its most likely source immediately. This urgency frequently arises before genomic sequencing is finished, before detailed contact networks are mapped, or before researchers can estimate a reliable, outbreak-specific serial-interval distribution. Documented person-to-person Andes virus (ANDV) transmission in Argentina, as well as recent outbreaks of mpox and Sudan virus disease, highlight this specific challenge~\citep{Akther2025NYCMpox,komakech2024svd,martinez2020super}. In every outbreak scenario, the core practical problem remains the same: based purely on the evidence available at that exact moment, which historical cases represent the most plausible sources of infection?

Existing reconstruction methods depend heavily on the type of data available. Genome-integrated frameworks like \texttt{outbreaker2}~\citep{campbell2018outbreaker2}, SCOTTI~\citep{de2016scotti}, and related epi-genomic approaches~\citep{didelot2014bayesian,didelot2017genomic} merge epidemiological timelines with pathogen sequencing. While these frameworks are highly effective, they require rich genomic data. When sequence data are missing or delayed, timing-based methods offer a simpler alternative. These tools typically fit Gaussian, Gamma, or Lognormal serial-interval distributions to score potential infectors. 

However, researchers rarely examine a critical limitation: the actual quality of the transmission labels used to train and test these models. Published parent--child links are widely accepted as definitive ground truth, even when they rely on subjective contact investigations, exposure histories, or incomplete genomic clues. Our separate audit of 134,095 Global.Health records highlights this severe shortage of benchmark-ready data. We identified only 26 recoverable transmission edges across the entire dataset. Furthermore, none of these edges provided a strict parent-ranking task backed by verified symptom onset dates~\citep{king2015avoidable,hadjisotiriou2023decision}.

This work examines both transferability and label reliability. A temporal transmission prior is learned from a multi-disease benchmark spanning eleven disease groups and then locked before any target-outbreak data are accessed. Its external performance is evaluated on a strict ANDV parent-ranking benchmark based on documented person-to-person transmission. No ANDV-specific refitting is performed. The study then examines transmission-label uncertainty using independent evidence from the 2022 New York City mpox outbreak and a densely traced Guangdong Delta outbreak. The mpox analysis tests whether epidemiological links are consistently supported by genomic evidence, while the Delta analysis tests whether retaining uncertain edges changes inferred source structure and fixed-capacity prioritization. These analyses do not assume the same degree of label uncertainty across all outbreaks. Rather, they test whether uncertainty in published transmission links can be large enough to affect reconstruction conclusions. Across the outbreaks examined here, uncertain links changed inferred source burden, source ranking, and the composition of prioritized cases.

Three forms of uncertainty help separate these questions. \emph{Inference uncertainty} concerns uncertainty within a reconstruction model over who infected whom. \emph{Label uncertainty} concerns uncertainty in the transmission links used as training or evaluation ground truth. \emph{Decision uncertainty} concerns changes in downstream conclusions, such as source ranking or intervention prioritization, that follow from the first two. Most reconstruction methods focus on inference uncertainty. This study focuses on the latter two: whether transmission labels used as ground truth are reliable, and whether uncertainty in those labels changes the decisions drawn from reconstructed outbreaks.

\section{Related Work}
\label{sec:2_related_works}

Transmission reconstruction methods generally follow two separate paths. The first path uses genomic sequence data to map transmission pathways, while the second relies purely on epidemiological timelines to score potential infectors. Our work sits between these two approaches. Crucially, we introduce a third dimension that neither path systematically covers: evaluating the reliability of the transmission labels that all methods ultimately depend on.

\paragraph{Parametric and nonparametric timing methods.}
\citet{WallingaTeunis2004} built the foundation for timing-based source attribution. They calculated probabilistic infector assignments using only case onset dates and a known serial interval distribution. This framework proved that temporal data alone provides a strong signal for identifying infectors. It sparked subsequent research into estimating generation intervals \citep{WallingaLipsitch2007, ganyani2020estimating} and tracking time-varying reproduction numbers \citep{Cori2013}. Documenting disease-specific serial intervals requires extensive fieldwork. For instance, the 14-year Nipah virus surveillance program in Bangladesh \citep{salje_nipah} highlights the massive epidemiological effort needed to get reliable interval data for just a single virus family. 

Parametric methods fit standard distributions such as Gaussian, Gamma, or Lognormal to observed interval data, scoring potential infector pairs by statistical likelihood. These models are highly interpretable, but choosing the wrong distribution introduces systematic bias if the real outbreak deviates from that specific shape. Nonparametric alternatives, such as kernel density estimators, avoid these shape assumptions entirely. However, they remain highly sensitive to bandwidth choices and struggle to predict gap values outside the sample data. Neither approach is explicitly trained to rank parents across different diseases. Instead, their performance relies heavily on the chosen density model and the source interval distribution. In contrast, our approach learns a discriminative timing score from source ranking labels, freezing it entirely before target evaluation.

\paragraph{Genome-integrated transmission reconstruction.}
A second group of methods merges pathogen sequences with epidemiological data to reconstruct transmission trees. \citet{didelot2014bayesian} formalized Bayesian inference for whole-genome sequence data. Later updates accounted for partially sampled or ongoing outbreaks, where unsequenced intermediate hosts create gaps in the data \citep{didelot2017genomic}. The \texttt{outbreaker2} platform \citep{campbell2018outbreaker2} provides a modular setup for combining epidemiological and evolutionary inference across diverse outbreak types. Similarly, the structured coalescent model in \texttt{SCOTTI} \citep{de2016scotti} treats each host as an independent subpopulation, explicitly modeling within-host diversity during transmission. Variant-aware models \citep{de2018bayesian} and epi-genomic frameworks \citep{kenah2016molecular, carson2025mbe} build on these ideas, using intrahost variant frequencies and genomic distances to sharpen transmission links. 

To handle larger datasets, \texttt{ScITree} \citep{scitree} introduces a highly scalable Bayesian framework for joint epidemiological and genomic inference. Recent preprint architectures, including JUNIPER \citep{juniper2025} and BREATH \citep{colijn2024breath}, showcase new directions in joint phylodynamic--epidemiological mapping. We note these purely as adjacent preprints rather than peer reviewed benchmarks. While genomic methods are incredibly powerful, they require high-quality whole-genome sequences with enough within-host diversity to isolate individual cases. They are not built for timing-only scenarios where genomic data are missing or delayed. As a result, direct numerical comparisons would be methodologically unfair. Our study acts as a complement to genome-integrated reconstruction, not a replacement for it.

\paragraph{Latent-variable inference under evidence uncertainty.}
A distinct methodological response to transmission uncertainty operates at the
inference layer rather than the benchmark layer. Instead of thresholding
phylogenetic evidence into fixed transmission calls, \citet{bu2024latent} model
HIV transmission linkage and direction as latent typed points and carry the
underlying phylogenetic linkage and direction scores through a fully Bayesian
data-augmented scheme, so that low-confidence pairs are down-weighted rather
than discarded. Such approaches address \emph{inference uncertainty} in the
sense defined in Section~\ref{sec:1_Intro}: they preserve evidential uncertainty
while estimating transmission structure on a single outbreak. The present work is complementary and occupies a different locus. Rather than modelling evidence uncertainty directly within inference, this framework treats published transmission links as inherently uncertain benchmarks. This approach allows for a precise quantification of how label uncertainty propagates into evaluation, inferred source structures, and downstream prioritization across outbreaks. Consequently, latent variable inference and the present label-audit framework address adjacent but non-overlapping layers of the taxonomy, functioning as complementary tools rather than competing methods.

\paragraph{Transmission benchmark construction and label reliability.}
Comparative evaluation of reconstruction methods requires ground truth labels linking each
case to its true infector. In practice, these labels derive from contact investigations,
phylogenetic proximity, or outbreak investigation reports; none constitutes direct
observation of a transmission event. The \texttt{OutbreakTrees} resource
\citep{taubeoutbreaktrees2022} curates a multi-disease collection of published transmission
trees that has enabled cross-method and cross-disease comparison, but label quality in
such resources is rarely formally characterized. \citet{king2015avoidable} documented how
errors in outbreak data structure propagate into reconstruction conclusions, motivating
explicit provenance auditing before benchmark construction. The present study extends this
concern in two ways. A systematic audit of 134,095 records from a large public
outbreak repository recovered only 26 transmission edges usable for parent-ranking under
strict construction criteria, illustrating the structural scarcity of benchmark-ready data.
A formal epi-genomic concordance audit of epidemiologically linked mpox pairs
\citep{Akther2025NYCMpox} demonstrates that label uncertainty is not merely a data-access
problem but a fundamental property of how transmission events are recorded. No prior work,
to our knowledge, has quantified the downstream consequence of this uncertainty for specific
public-health prioritization decisions: specifically, whether choosing between strict and
uncertainty-aware graph construction changes which source cases are selected under
fixed-capacity response scenarios.

\paragraph{Robustness diagnostics and uncertainty-aware decision analysis.}
Statistical conclusions from compact outbreak benchmarks are vulnerable to the
influence of individual cases. Robustness diagnostics for binary trial endpoints, developed
under the fragility index framework \citep{Walsh2014}, quantify how many outcome reversals
would overturn a significance conclusion. The present study adapts this concept to the
paired ranking setting, computing the minimum number of task-level reversals required to
lose sign-test significance for each method comparison. This paired reversal index differs
from the classical fragility index, which targets dichotomous trial outcomes, but shares
its interpretive advantage: robustness expressed in units directly meaningful to the
evaluation design. Alongside resampling-based inference \citep{WassersteinLazar2016} and
leave-one-out influence diagnostics, this framework provides multidimensional robustness
characterization suited to compact real-outbreak benchmarks. At the decision level,
\citet{hadjisotiriou2023decision} studied the consequences of graph-level uncertainty for
policy prioritization under deep uncertainty frameworks, arguing for stress-testing
conclusions across plausible alternative scenarios rather than optimizing against a single
point forecast. The present study operationalizes this principle in a concrete outbreak
context by comparing strict and uncertainty-expanded transmission graphs and measuring
priority-set instability under fixed response capacity. The decision curve analysis
framework \citep{VickersElkin2006} provides additional conceptual basis for evaluating
the operational consequences of threshold-based classification decisions under uncertainty. Table~\ref{tab:related_work_comparison} positions the proposed method against
representative timing-based, genome-integrated, and uncertainty-aware
transmission-reconstruction approaches.

\begin{table*}[t]
\centering
\caption{Representative transmission-reconstruction approaches and their
relation to the present study.}
\label{tab:related_work_comparison}

\small
\setlength{\tabcolsep}{5pt}
\renewcommand{\arraystretch}{1.18}

\begin{tabularx}{\textwidth}{
>{\raggedright\arraybackslash}p{3.0cm}
>{\raggedright\arraybackslash}p{3.0cm}
>{\raggedright\arraybackslash}p{3.0cm}
>{\centering\arraybackslash}p{1.5cm}
>{\centering\arraybackslash}p{1.8cm}
>{\raggedright\arraybackslash}X
}
\toprule

\textbf{Approach} &
\textbf{Primary input} &
\textbf{Main output} &
\textbf{Genomics} &
\textbf{Label uncertainty} &
\textbf{Role in this study} \\

\midrule

\textbf{Locked temporal prior}
\newline \emph{This study}
&
Onset-time gaps
&
Candidate-parent ranking without target refitting
&
$\times$
&
\textit{Separate audit}
&
Zero-shot temporal ranking; label uncertainty evaluated separately
\\

\textbf{Serial-interval priors}
\newline Gaussian, Gamma, Lognormal, KDE
&
Observed transmission intervals
&
Temporal likelihood for candidate infectors
&
$\times$
&
--
&
Fair timing-only baselines
\citep{WallingaTeunis2004,WallingaLipsitch2007,ganyani2020estimating}
\\

\textbf{Wallinga--Teunis}
&
Case onset times + serial interval
&
Probabilistic source attribution
&
$\times$
&
--
&
Foundational timing-only attribution
\citep{WallingaTeunis2004}
\\

\textbf{\texttt{outbreaker2}}
&
Timing, contacts, pathogen sequences
&
Posterior transmission trees
&
\checkmark
&
--
&
Genome-integrated reconstruction
\citep{campbell2018outbreaker2,didelot2014bayesian}
\\

\textbf{\texttt{SCOTTI}}
&
Multiple sequences per host + sampling times
&
Host-to-host transmission under a structured coalescent
&
\checkmark
&
--
&
Models within-host evolution and unsampled intermediates
\citep{de2016scotti}
\\

\textbf{Variant-aware / partially sampled epi-genomic methods}
&
Genomes, variants, timing, epidemiological links
&
Transmission-tree inference with incomplete sampling
&
\checkmark
&
--
&
Richer inference when genomic evidence is available
\citep{de2018bayesian,didelot2017genomic,kenah2016molecular,carson2025mbe}
\\

\textbf{\texttt{ScITree}}
&
Epidemiological + genomic data
&
Scalable posterior transmission-tree inference
&
\checkmark
&
--
&
Large-outbreak genome-integrated inference
\citep{scitree}
\\

\textbf{Latent evidence models}
&
Phylogenetic linkage/direction scores
&
Transmission inference with uncertain evidence states
&
\checkmark
&
\textit{Inference-level}
&
Addresses inference-level evidence uncertainty
\citep{bu2024latent}
\\

\textbf{JUNIPER / BREATH}
&
Dense sequencing + epidemiological information
&
Joint phylogenetic/transmission histories
&
\checkmark
&
--
&
Emerging data-rich reconstruction frameworks
\citep{juniper2025,colijn2024breath}
\\

\textbf{Decision / robustness analysis}
&
Alternative graph or outcome scenarios
&
Priority instability and robustness
&
$\times$
&
\textit{Decision-level}
&
Conceptual basis for downstream uncertainty analysis
\citep{hadjisotiriou2023decision,Walsh2014,WassersteinLazar2016}
\\

\bottomrule
\end{tabularx}
\end{table*}

\section{Methods}
\label{sec:3_methods}

Figure~\ref{fig:fig_1_study_design} summarizes the study design. Initially, a temporal transmission prior was trained on the D1 multi-disease benchmark and locked before accessing any target-outbreak data. This fixed prior was subsequently evaluated on a strict Andes virus (ANDV) parent-ranking benchmark. Finally, independent analyses of MPXV and Guangdong Delta datasets examined transmission-label uncertainty alongside its downstream impact on source prioritization.

\begin{figure}[t]
    \centering
    \includegraphics[width=\textwidth]{Figure_1_study_design_evidence_architecture.png}
    \caption{\textbf{Study design and evidence architecture.}
    The temporal prior is learned from D1 and evaluated without target-specific
    refitting on the strict ANDV benchmark. The MPXV and Guangdong Delta
    modules independently assess label reliability and graph-level uncertainty.}
    \label{fig:fig_1_study_design}
\end{figure}

% ============================================================
\subsection{Candidate-Infector Ranking}
\label{sec:3_1_problem_formulation}

For each target case $j$, the candidate set $\mathcal{C}(j)$ includes all cases with documented onset dates falling within a predefined temporal window before $j$'s onset. In our strict benchmark, exactly one candidate represents the true documented parent $p_j$. Our primary objective is to rank $p_j$ as highly as possible among these eligible candidates.

We use Mean Reciprocal Rank (MRR) as our primary evaluation metric:
\begin{equation}
\label{eq:mrr}
\mathrm{MRR}
=
\frac{1}{|T|}
\sum_{j\in T}
\frac{1}{\mathrm{rank}(p_j)},
\end{equation}
where $T$ represents the total set of ranking tasks. Additionally, we report Top-1 accuracy and true-parent rank as secondary summaries.

To build the strict external benchmarks, we applied a prespecified temporal eligibility rule. Specifically, the documented parent had to be unique, possess a verified onset date, fall within the external benchmark window, and carry high-confidence evidence. Algorithm~\ref{alg:1_strict_candidate_benchmark} outlines this entire setup. Finally, we constructed the D1 source benchmark separately from curated transmission-tree candidate sets, following the process detailed in Section~\ref{sec:3_2_Multi_Disease_benchmark}.

\begin{algorithm}[t]
\caption{Strict external candidate-infector benchmark construction.}
\label{alg:1_strict_candidate_benchmark}
\KwIn{Cases $\mathcal{V}$, directed evidence edges $\mathcal{E}$,
edge confidence $q(e)$, and temporal window $[w_{\min},w_{\max}]$.}
\KwOut{Ranking tasks $\mathcal{B}=\{(j,\mathcal{C}_j,p_j)\}$.}

Initialize $\mathcal{B}\leftarrow\emptyset$\;

\ForEach{target case $j$}{
    require one documented high-confidence parent $p_j$\;
    require verified onset dates for $j$ and $p_j$\;
    require $w_{\min}\leq d(j)-d(p_j)\leq w_{\max}$\;
    set $\mathcal{C}_j$ to all cases $i\neq j$ satisfying
    $w_{\min}\leq d(j)-d(i)\leq w_{\max}$\;
    \If{$p_j\in\mathcal{C}_j$}{
        add $(j,\mathcal{C}_j,p_j)$ to $\mathcal{B}$\;
    }
}
\Return $\mathcal{B}$\;
\end{algorithm}

% ============================================================
\subsection{D1 Multi-Disease Training Benchmark}
\label{sec:3_2_Multi_Disease_benchmark}

\paragraph{Source data.}
\label{sec:3_2_1_source_data}

D1 was constructed from published transmission trees spanning eleven disease
groups. Each tree contained directed parent--child links and associated symptom
onset dates. The final benchmark included COVID-19, MERS, Nipah virus~\citep{salje_nipah},
Smallpox, Ebola, combined H1N1/H3N2 influenza, Pneumonic plague, Measles,
Rubella, Norovirus, and H7N9.

\paragraph{Benchmark construction.}
\label{sec:3_2_2_benchmark_construction}

D1 candidate sets were inherited from the curated multi-disease
transmission-tree benchmark and its frozen data split. Timing correction and
quality control were then applied to these candidate rows, and the signed
onset gap between each candidate source and target was calculated. The ANDV
1--60day predecessor rule was not reapplied to D1. Consequently, D1 contains
negative, zero, and positive onset gaps, reflecting that symptom-onset order
need not always coincide with transmission order. Records with missing timing information, self-loops, or non-unique strict
positives were excluded. The final D1 benchmark contained 14,919 candidate
rows, 559 true-parent rows, 559 ranking tasks, and 11 disease groups. Every
retained task contained exactly one documented parent.

\paragraph{Leave-one-disease-out evaluation.}
\label{sec:3_2_3_lodo}

Cross-disease generalization was evaluated using leave-one-disease-out (LODO)
validation. In each fold, one disease group was excluded from training and used
only for evaluation. Disease-level MRR and Top-1 accuracy were calculated for
each fold, and disease-macro estimates were calculated by averaging across the
11 held-out groups. Confidence intervals were estimated from 10,000 bootstrap
resamples of the disease folds.

% ============================================================
\subsection{Learned Temporal Prior}
\label{sec:3_3_learned_temporal_prior}

\paragraph{Feature representation.}
\label{sec:3_3_1_feature_representation}

For candidate parent $i$ and target case $j$, the basic input is the onset gap
\begin{equation}
\label{eq:serial_gap}
\Delta t = \mathrm{onset}(j)-\mathrm{onset}(i).
\end{equation}
The timing-only feature vector contains $\Delta t$, $|\Delta t|$, an indicator
for $\Delta t\geq0$, indicators for negative, zero, and positive gaps,
$(\Delta t)^2$, and $(|\Delta t|)^2$. No demographic, clinical, spatial,
contact-network, or genomic variables were used. Candidate eligibility was
defined separately by the 1--60day window.

\paragraph{Model.}
\label{sec:3_3_2_model_architecture}

A logistic regression model was trained on the binary parent label $y\in\{0,1\}$, utilizing eight timing features standardized via parameters estimated directly from the source training data. This model specification utilized $C=0.01$, \texttt{balanced} class weighting, the \texttt{lbfgs} solver, a maximum of 5,000 iterations, and a fixed random seed of 20260514. 
The resulting model output,
\[
P(\Delta t)=
\sigma\!\left(\hat{\theta}^{\top}\phi(i,j)\right),
\]
served exclusively as a relative ranking score within each candidate set, rather than functioning as a calibrated marginal probability. For the Leave-One-Disease-Out (LODO) evaluation, the preprocessing and model fitting steps were repeated using only the source diseases available within each respective fold. Once the final model specification was fixed, the external locked prior was fitted solely to the predefined D1 training split, which comprised 14,053 candidate rows, 466 true-parent rows, and 38 transmission trees.

\paragraph{Locking protocol.}
\label{sec:3_3_3_Prior_locking_external_validation}

Following source training, the model parameters were frozen and stored as a serialized artifact alongside the temporal plausibility curve. External benchmarks were subsequently scored using this fixed artifact, bypassing any refitting, recalibration, or target-specific parameter adjustments. Reapplying this stored artifact to the ANDV benchmark successfully reproduced the archived ranking outputs.

\paragraph{Prior-shape analysis.}
\label{sec:3_3_4_Temporal_prior_shape}

The locked temporal curve was visualized after applying peak normalization. To assess architectural stability across different diseases, pairwise Pearson and Spearman correlations were calculated between the 11 LODO-trained temporal curves.

% ============================================================
\subsection{Temporal Baselines}
\label{sec:3_4_baselines_comparators}

Four source-trained, temporal-likelihood baselines were evaluated under the same target-free protocol. Specifically, a Gaussian density was fitted to the observed source gaps, and Scott's bandwidth rule was applied to a Gaussian kernel density estimator. Additionally, Gamma and Lognormal distributions were fitted to positive gaps, assigning a zero score to any non-positive gaps. Estimation for all four baselines relied exclusively on the D1 source data. An Epuy\'en-specific Gaussian serial-interval reference (with a mean of approximately 23 days and a standard deviation of roughly 7 days) was included as an informed contextual comparator. Because this reference relies on ANDV-specific timing information, it was not treated as a fair, source-trained baseline.

% ============================================================
\subsection{Real ANDV Benchmark}
\label{sec:3_5_andv_benchmark_construction}

\paragraph{Source outbreaks.}
\label{sec:3_5_1_source_outbreaks}

The primary external benchmark combines data from two published person-to-person ANDV outbreak investigations. The first, the 2018--2019 Epuy\'en outbreak, contains 34 confirmed cases spanning four transmission generations \citep{martinez2020super}. Case-level onset dates and transmission links were extracted directly from the main report and Supplementary Table S2. The second, a separate three-case 2014 Argentina transmission chain, provided additional documented onset times and direction data. Because ANDV was completely absent from the D1 dataset, this evaluation serves as an external zero-shot test.

\paragraph{Strict edge criteria.}
\label{sec:3_5_2_strict_edge_inclusion_criteria}

A parent--child pair was retained only when:
\begin{enumerate}
    \item both onset dates were documented and not imputed;
    \item the parent--child gap was 1--60 days;
    \item the source report classified the link as high-confidence or
    confirmed; and
    \item the child had no competing high-confidence parent.
\end{enumerate}

Uncertain, possible, and alternative links were excluded from the primary
benchmark and retained only for the uncertainty analysis. The resulting
strict benchmark contained 37 cases, 29 true parent--child edges,
29 ranking tasks, and 395 candidate-parent rows. Twenty-seven tasks contained
at least two candidate parents.

\paragraph{Candidate sets.}
\label{sec:3_5_3_candidate_set_construction}

For target case $j$, all cases with documented onset dates in
\begin{equation}
\label{eq:candidate_window}
[\,\mathrm{onset}(j)-60,\;\mathrm{onset}(j)-1\,]
\end{equation}
were eligible candidates. Two tasks contained only one eligible predecessor.
They were retained in the primary benchmark for completeness but contribute
equally to all methods. Sensitivity analyses therefore also evaluated the
27 non-singleton tasks.

% ============================================================
\subsection{Secondary Benchmark-Readiness Analyses}

\paragraph{Global.Health audit.}
\label{sec:3_6_globalhealth_repo_audit}

The Global.Health line-list repository was screened for information needed
to construct strict ranking tasks: verified onset dates, directed parent links,
edge confidence, and unique-parent structure. Among 134,095 records, only
26 recoverable transmission edges satisfied the strict linkage criteria and
none yielded a ranking task with verified parent and child onset dates. This
analysis assessed benchmark readiness rather than model performance.

\paragraph{Sudan virus disease pilot.}
\label{sec:3_7_svd_pilot_benchmark}
\label{sec:3_7_1_network_reconstruction}
\label{sec:3_7_2_Pilot_Benchmark_construction}

A small exploratory benchmark was reconstructed from published records of the
2022 Uganda Sudan virus disease outbreak
\citep{komakech2024svd,kabami2024svd}. Absolute onset dates were not uniformly
available, so relative temporal positions were used instead. Candidate-parent
ranking was evaluated using simple gap-proximity heuristics, including a
14day reference rule and a shortest-gap rule. The locked temporal prior was
not evaluated under its external zero-shot protocol on this dataset; the SVD
analysis therefore provides only relative-time feasibility evidence.

% ============================================================
\subsection{Statistical Evaluation and Sensitivity Analyses}
\label{sec:3_8_statistical_evo_robustness}

\paragraph{Primary comparisons.}
\label{sec:3_8_1_primary_statistical_test}

For each ANDV task, the learned prior was compared with each temporal baseline
using the reciprocal-rank difference
\begin{equation}
\label{eq:delta_mrr}
\Delta\mathrm{MRR}(k)
=
\mathrm{MRR}_{\mathrm{learned}}(k)
-
\mathrm{MRR}_{\mathrm{baseline}}(k).
\end{equation}
Mean $\Delta\mathrm{MRR}$ was the primary effect size. Uncertainty was evaluated
with 10,000 task-level bootstrap resamples. One-sided sign-flip permutation
tests used 100,000 repetitions, and one-sided Wilcoxon signed-rank tests
assessed rank differences. Top-1 discordance was tested with an exact binomial
test. Benjamini--Hochberg correction was applied across the four fair baseline
comparisons for confirmatory reporting.

\paragraph{Finite-sample robustness.}
\label{sec:3_8_2_finite_sample_robustness}

Sensitivity to individual ANDV tasks was assessed with leave-one-task-out
analysis and jackknife standard errors. A paired task-reversal index measured
the minimum number of learned-better tasks that would need to be reversed before
the one-sided sign test became non-significant at $\alpha=0.05$. Candidate
windows of 1--30, 1--45, 1--60, 4--45, 4--60, 7--45, and 7--60 days were also
evaluated.

\paragraph{Source-domain and negative-control analyses.}
Source-domain sensitivity was evaluated under three targeted retraining
conditions: removal of MERS, removal of Smallpox, and removal of both groups.
Smallpox was included as a targeted control because it was the only
orthopoxvirus group in D1, allowing us to test whether the ANDV result depended
disproportionately on that distinct source-domain timing profile rather than on
the broader multi-disease temporal structure. Each model used the same
preprocessing and training protocol as the locked prior and was evaluated on
the unchanged 29-task strict ANDV benchmark. As a negative control, D1 parent
labels were randomly permuted 2,000 times while preserving the feature matrix,
candidate structure, and training procedure. These analyses were diagnostic
and were not used to tune or select the primary model.

% ============================================================
\subsection{MPXV Transmission-Label Audit}
\label{sec:3_9_mpxv_level_uncertainty}

\paragraph{Data and classification.}
\label{sec:3_9_1_source_data}
\label{sec:3_9_2_codebook_classification}
\label{sec:3_9_3_codebook_reproducibility}

Transmission-label reliability was examined using the 2022 New York City mpox
genomic epidemiology study of Akther et al.\
\citep{Akther2025NYCMpox}. The source study sequenced 1,138 MPXV genomes and
reported phylogenetic concordance for epidemiologically linked cases. After
excluding within-host comparisons, 75 inter-host linked pairs were retained. Each published concordance label was deterministically mapped to one of four
evidence classes:
\begin{description}
    \item[Strict supported:] an exclusive monophyletic relationship consistent
    with direct transmission;
    \item[Potential supported:] a recent shared ancestor without an exclusive
    clade;
    \item[Unresolved:] insufficient phylogenetic resolution; and
    \item[Not supported:] genomic distance inconsistent with direct
    transmission.
\end{description}
The mapping was fixed before aggregate analysis and reproduced the published
concordance counts. Because each published concordance label mapped deterministically to a single
evidence class, no independent manual coding or inter-rater adjudication was
performed; inter-rater agreement statistics such as Cohen's $\kappa$ were
therefore not applicable.

\paragraph{Statistical analysis.}
\label{sec:3_9_4_statistical_analysis}

The primary quantity was the proportion of pairs classified as unresolved or
not supported. An exact binomial confidence interval was calculated for this
proportion. This module assesses transmission-label reliability; it is
independent of the temporal-prior validation.

% ============================================================
\subsection{Guangdong Delta Transmission Graph}
\label{sec:3_10_guangdong_delta_graph}

\paragraph{Graph extraction.}
\label{sec:3_10_1_source_data}

Graph-level uncertainty was examined using the published transmission
visualization from the 2021 Guangdong SARS-CoV-2 Delta outbreak
\citep{li2022viral}. Nodes and directed edges were obtained from
\texttt{cases.json} and \texttt{case\_connections\_raw.json}, preserving the
source distinction between high-confidence solid edges and uncertain dashed
edges. The extracted visualization contained 131 nodes and 142 directed edges,
including 107 high-confidence and 35 uncertain edges. Among human
case-to-case links, 57 met the strict unique-parent criterion and five child
cases had multiple high-confidence possible parents.

\paragraph{Analytical scope.}
\label{sec:3_10_2_analytical_role_limitation}

The available visualization files did not contain a defensible case-level
symptom-onset field for constructing a temporal parent-ranking benchmark.
The Delta data were therefore used only for graph-level uncertainty analysis,
not for evaluation of the learned temporal prior.

% ============================================================
\subsection{Transmission-Label Uncertainty and Decision Instability}
\label{sec:3_11_transmission_label_uncertainty}

\paragraph{Graph expansion.}
\label{sec:3_11_1_uncertainty_Expansion}

For ANDV, three edge sets were considered: strict high-confidence links,
a preferred set adding lower-confidence alternatives with a preferred parent,
and a full plausible set containing all epidemiologically plausible links.
For Guangdong Delta, the high-confidence graph was compared with a graph that
also retained uncertain dashed-line case-to-case edges. Offspring counts and
source rankings were recomputed under each edge set.

\paragraph{Source-ranking stability.}
\label{sec:3_11_2_source_ranking}

Structural changes were summarized using top-$k$ source-set Jaccard similarity,
Spearman and Kendall rank correlations, and the Gini coefficient of the
offspring-count distribution. Together, these measures capture changes in the
identity, ordering, and concentration of inferred source burden.

\paragraph{Decision instability.}
\label{sec:3_11_3_Decision_instability}

A fixed-capacity scenario was used to assess whether graph uncertainty changes
which source cases would be prioritized for investigation. The primary
comparison used the top five source cases under the strict and expanded
graphs. Priority-set overlap was measured by Jaccard similarity, and cases
crossing an offspring threshold of three were recorded as newly elevated
sources.

Decision regret was defined as the fraction of expanded-graph top-5 offspring
burden not captured by strict-only prioritization:
\begin{equation}
\label{eq:decision_regret}
\mathrm{Regret}
=
1-
\frac{
\displaystyle
\sum_{i\in S_{\mathrm{strict}}\cap S_{\mathrm{expanded}}}
\mathrm{offspring}_{\mathrm{expanded}}(i)
}{
\displaystyle
\sum_{i\in S_{\mathrm{expanded}}}
\mathrm{offspring}_{\mathrm{expanded}}(i)
}.
\end{equation}
This directional measure quantifies sensitivity to retained edge uncertainty;
it does not assume that the expanded graph is the true transmission network or
that expanded-graph prioritization would improve intervention outcomes.

% ============================================================

\section{Results}
\label{sec:results}

\subsection{Cross-Disease Transfer and Prior Stability}
\label{sec:4_2_cross_disease_results_d1}

Our learned temporal prior generalized effectively across all 11 D1 disease groups. Under a leave-one-disease-out (LODO) evaluation, the model achieved a disease-macro MRR of 0.57495. In comparison, the source-trained Gaussian baseline yielded an MRR of only 0.45225, demonstrating a clear improvement ($\Delta\mathrm{MRR}=+0.12270$; bootstrap 95\% CI $[+0.02978,+0.25395]$). Furthermore, the learned prior outperformed the baseline in 7 out of the 8 discordant disease folds (one-sided exact sign test, $p=0.0352$). 

\begin{figure}[t]
\centering
\includegraphics[width=\linewidth]{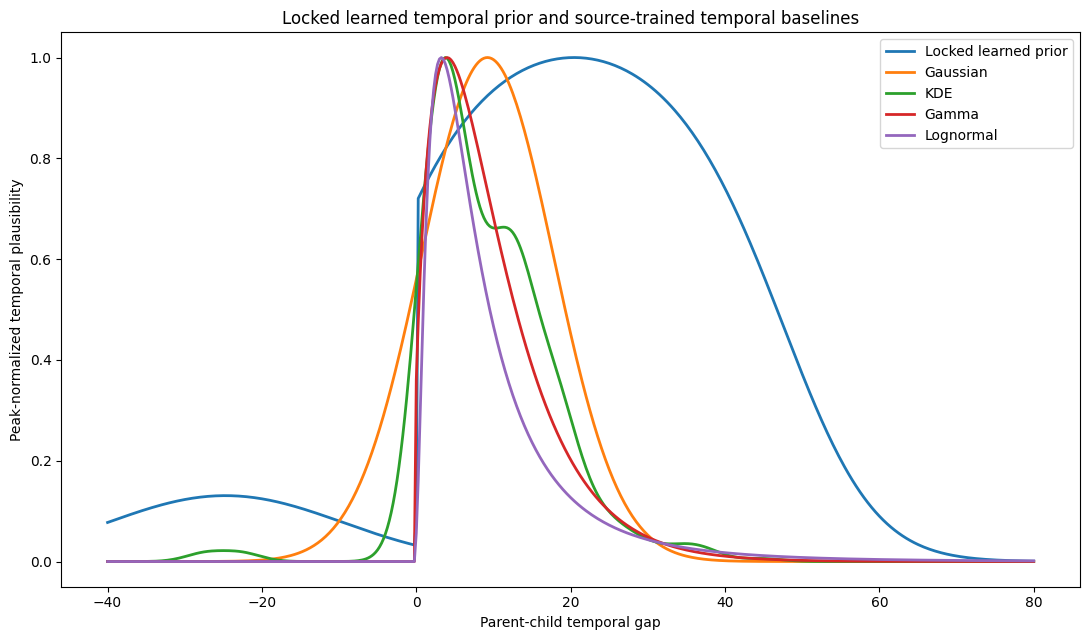}
\caption{Peak-normalized D1 temporal prior and source-trained baselines.}
\label{fig:fig_2_temporal_prior_curves}
\end{figure}

The underlying temporal shape also remained highly stable across held-out diseases. Across 55 pairwise comparisons of LODO-trained curves, we observed a mean Pearson correlation of 0.9917 and a mean Spearman correlation of 0.9694. Introducing higher-capacity listwise models failed to improve disease-macro performance, while few-shot adaptation offered only negligible gains. Together, these findings strongly support using a compact, source-trained prior that transfers cleanly without requiring target-specific refitting. We plot the locked temporal prior alongside the four source-trained temporal baseline curves in Figure~\ref{fig:fig_2_temporal_prior_curves}.

\begin{figure}[t]
\centering
\includegraphics[width=\linewidth]{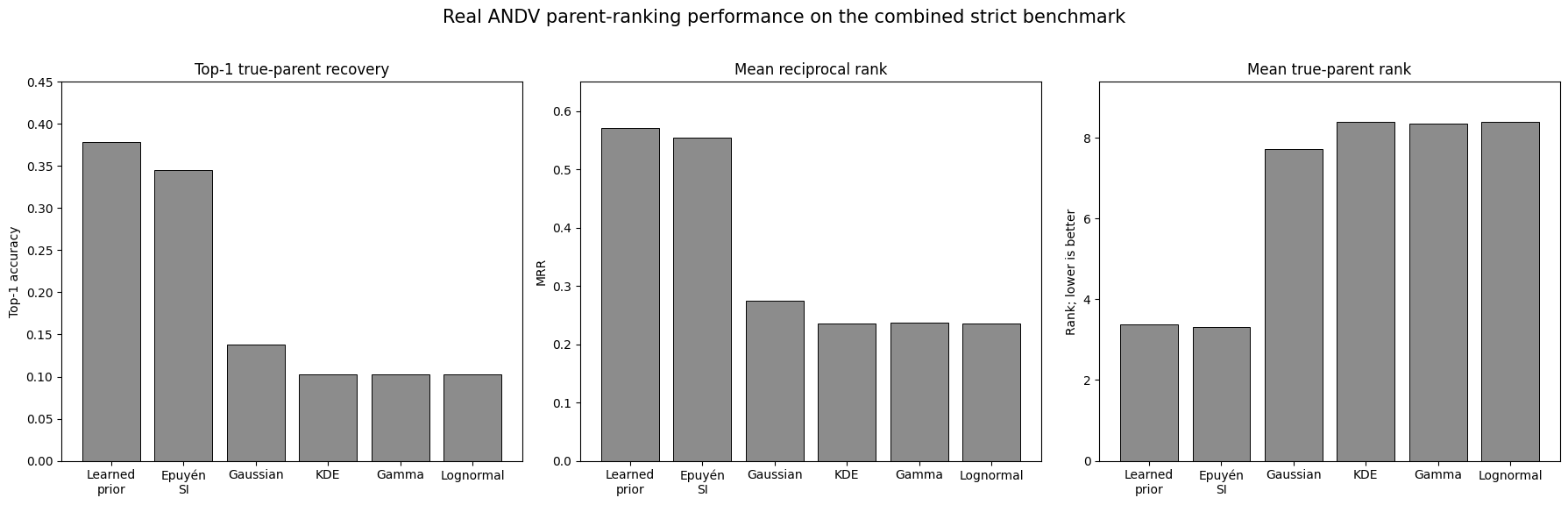}
\caption{Parent-ranking performance on the strict ANDV benchmark ($n=29$ tasks).}
\label{fig:fig_3_ANDV_primary_parent_ranking_performance}
\end{figure}

% ============================================================
\subsection{External Validation on Real ANDV}
\label{sec:results_andv_performance}

The strict ANDV benchmark contained 29 parent-ranking tasks, of which 27 had
at least two eligible candidate parents. The locked prior achieved Top-1
accuracy of 0.379 and MRR of 0.571, compared with 0.138 and 0.274,
respectively, for the strongest fair source-trained baseline (Gaussian).
Median true-parent rank improved from 7 under the Gaussian baseline to 2 under
the learned prior. KDE, Gamma, and Lognormal likelihoods performed similarly,
with MRR between 0.236 and 0.237. The outbreak-specific Epuy\'en
serial-interval reference achieved MRR 0.555 and Top-1 accuracy 0.345. Thus,
the source-trained learned prior substantially outperformed all fair
source-trained temporal baselines and performed comparably to a reference
calibrated with ANDV-specific timing information.

Paired task-level analysis confirmed that the improvement did not result from
aggregate averaging. Relative to Gaussian, mean MRR increased by 0.297
(95\% CI [0.156, 0.430]), and mean true-parent rank improved by 4.345
positions. The MRR permutation test gave $p=0.0002$, while the rank-based
Wilcoxon test gave $p=0.00007$. The same pattern held against KDE, Gamma, and
Lognormal, with $\Delta\mathrm{MRR}=0.334$--$0.335$ and permutation
$p\leq 0.00002$. In contrast, the difference from the Epuy\'en-specific
reference was small ($\Delta\mathrm{MRR}=0.016$) and not statistically
significant ($p=0.403$). Figure~\ref{fig:fig_3_ANDV_primary_parent_ranking_performance}
summarizes parent-ranking performance on ANDV, and
Table~\ref{tab:andv_primary_performance} reports the primary values and
MRR confidence intervals.

\begin{table}[t]
\centering
\caption{Parent-ranking performance on the strict ANDV benchmark
($n=29$ tasks). The Epuy\'en SI reference uses ANDV-specific timing
information and is included for context.}
\label{tab:andv_primary_performance}

\small
\setlength{\tabcolsep}{5.5pt}
\renewcommand{\arraystretch}{1.12}

\begin{tabular}{lccc}
\toprule
\textbf{Method} &
\textbf{Top-1} &
\textbf{MRR (95\% CI)} &
\textbf{Median Rank} \\
\midrule

Locked learned prior
& 0.379
& 0.571 [0.444, 0.701]
& 2 \\

Epuy\'en SI reference
& 0.345
& 0.555 [0.432, 0.680]
& 2 \\

Gaussian
& 0.138
& 0.274 [0.172, 0.393]
& 7 \\

KDE
& 0.103
& 0.236 [0.147, 0.347]
& 8 \\

Gamma
& 0.103
& 0.237 [0.147, 0.347]
& 8 \\

Lognormal
& 0.103
& 0.236 [0.147, 0.347]
& 8 \\

\bottomrule
\end{tabular}
\end{table}

% ============================================================
\subsection{Robustness and Sensitivity}
\label{sec:results_robustness}
\label{sec:4_9_candidate_andv_window_sensitivity}

Our ANDV findings remained stable despite the small size of the benchmark dataset. The bootstrap 95\% CI for the learned-prior MRR was [0.444, 0.701], while the jackknife standard error was 0.067. Notably, removing any single task shifted the MRR by a maximum of only 0.0187. For the performance gap to lose its one-sided sign-test significance, seven tasks where the prior performed better would need to reverse against the Gaussian baseline. For the KDE, Gamma, and Lognormal baselines, eight reversals were required to erase this significance. 

This conclusion was also highly insensitive to how we defined the candidate-window. Across seven alternative windows spanning from 1--30 to 7--60 days, the learned prior consistently remained the strongest fair timing-only method. Within these variations, its MRR ranged between 0.571 and 0.607, maintaining a positive advantage over the Gaussian baseline in every single window. Dropping the two singleton tasks yielded the same qualitative outcome: on the remaining 27 nontrivial tasks under our reference 1--60day window, the learned prior achieved an MRR of 0.539, compared with 0.211
for the Gaussian model.

Targeted source-domain ablations showed that the ANDV result depended partly
on the composition of D1 but was not explained by a single obvious source
analogue. ANDV MRR was 0.571 with the full D1 source training set, 0.507
after removing Smallpox, 0.483 after removing MERS, and 0.571 after removing
both groups. In the 2,000-label-permutation negative control, randomized
source labels produced mean MRR 0.318; the observed locked-prior result lay
in the upper tail of this null distribution (empirical $p=0.0395$). 

Finally, we conducted two secondary analyses to evaluate broader data availability rather than our primary transfer claim. First, screening 134,095 Global.Health records yielded only 26 recoverable transmission links and zero strict parent-ranking tasks with verified parent and child onset dates. Second, a small Sudan virus disease analysis indicated that relative temporal proximity can provide useful ranking information. However, this secondary test did not evaluate our locked prior under the strict external-validation protocol. Consequently, neither of these secondary analyses alters our primary external validation findings.

% ============================================================
\subsection{Transmission Labels Contain Substantial Uncertainty}
\label{sec:4_9_results_mpxv_evidence}
\label{sec:results_delta}

Our independent MPXV audit examined 75 epidemiologically linked inter-host pairs to evaluate baseline label reliability. Among these, 21 pairs possessed strict genomic support, 13 had potential support, 31 were unresolved, and 10 completely lacked genomic backing. Combining the unresolved and unsupported categories revealed that 41 out of 75 pairs (54.67\%; exact 95\% CI: 42.75--66.21\%) could not be classified as confirmed direct-transmission links. Because this confidence interval spans 0.50, we interpret these findings as
evidence of substantial label uncertainty in the audited pair set, rather
than evidence that the underlying unresolved-or-unsupported proportion
exceeds one-half.

Our analysis of the Guangdong Delta transmission graph revealed a different pattern of uncertainty. The extracted network contained 35 uncertain directed edges, and five child cases had multiple high-confidence candidate parents. We utilized these specific observations solely to measure graph-level uncertainty. Crucially, the public visualization lacked the defensible onset dates required for temporal parent-ranking, making further modeling inappropriate.

% ============================================================
\subsection{Uncertainty Changed Source Structure and Prioritization}
\label{sec:results_uncertainty_structure}
\label{sec:results_decision_instability}
\label{sec:results_threshold_shifts}

Retaining plausible uncertain links changed source structure in both outbreak
graphs. In Epuy\'en, the graph expanded from 27 strict edges to 40 plausible
edges; nine source cases gained possible offspring. In Guangdong Delta, the
corresponding expansion was from 57 strict unique-parent edges to 72
high-confidence-plus-uncertain case links, with 11 parent nodes gaining
offspring. These changes altered which sources ranked highest. Top-5 source-set Jaccard
similarity between strict and expanded graphs was 0.667 for ANDV and 0.429 for
Delta. Offspring-burden concentration also decreased: the Gini coefficient
shifted from 0.691 to 0.453 in ANDV and from 0.500 to 0.374 in Delta. Under a top-5 investigation capacity, strict-only prioritization missed P22 in
ANDV and source nodes 5646 and 5647 in Delta. The corresponding decision-regret
fractions were 0.0714 and 0.0357. These values are moderate; the important
change is in the identity of the prioritized cases rather than a large loss of
aggregate offspring coverage. The same pattern appeared under the high-priority threshold of at least three
offspring. The number of qualifying sources increased from 3 to 5 in ANDV,
with P10 and P22 newly elevated, and from 7 to 8 in Delta, with source 5646
newly elevated.

\begin{table}[t]
\centering
\caption{Effect of retaining uncertain transmission links on inferred source
structure and fixed-capacity prioritization.}
\label{tab:uncertainty_source_summary}
\label{tab:graph_uncertainty_source_structure}
\label{tab:decision_instability}
\label{tab:high_priority_source_shifts}

\small
\setlength{\tabcolsep}{5pt}
\renewcommand{\arraystretch}{1.15}

\begin{tabularx}{\linewidth}{
>{\raggedright\arraybackslash}p{2.2cm}
>{\centering\arraybackslash}p{2.0cm}
>{\centering\arraybackslash}p{1.5cm}
>{\centering\arraybackslash}p{2.1cm}
>{\centering\arraybackslash}p{1.5cm}
>{\raggedright\arraybackslash}X
}
\toprule
\textbf{Dataset} &
\textbf{Edges} &
\makecell{\textbf{Top-5}\\\textbf{Jaccard}} &
\textbf{Gini Shift} &
\makecell{\textbf{Decision}\\\textbf{Regret}} &
\textbf{New High-Priority Sources} \\
\midrule
ANDV Epuy\'en
&
$27 \rightarrow 40$
&
0.667
&
$0.691 \rightarrow 0.453$
&
0.0714
&
P10, P22
\\

Guangdong Delta
&
$57 \rightarrow 72$
&
0.429
&
$0.500 \rightarrow 0.374$
&
0.0357
&
5646
\\
\bottomrule
\end{tabularx}
\end{table}

\begin{figure}[t]
    \centering
    \includegraphics[width=\textwidth]{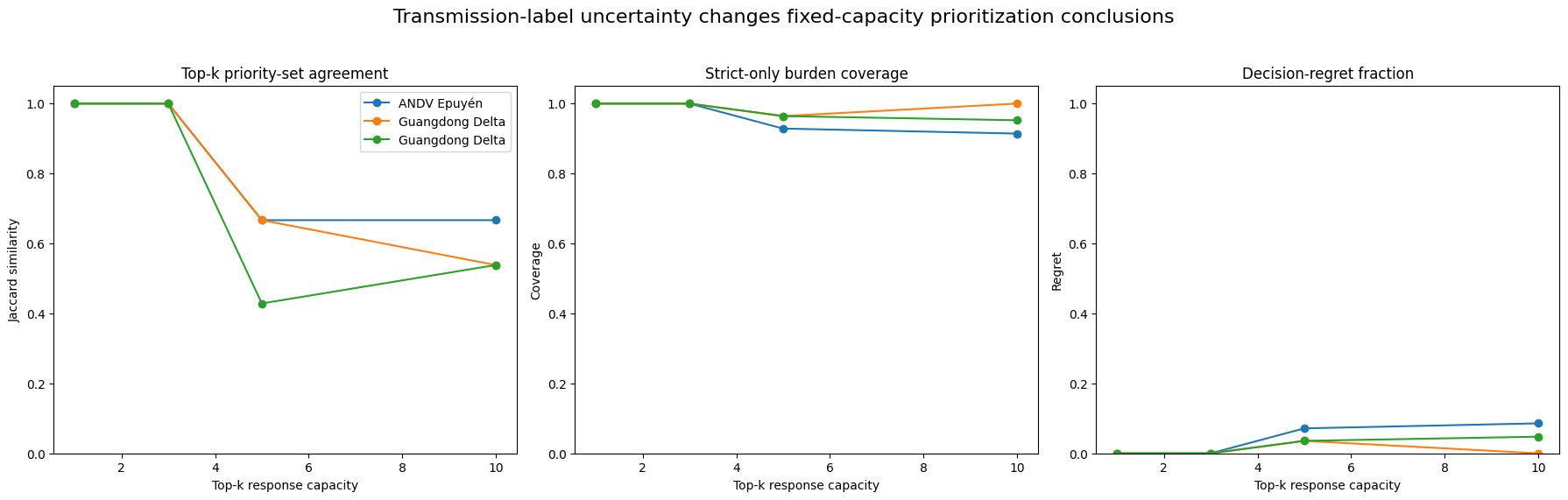}
    \caption{\textbf{Priority-set stability across response capacities.}
    Comparison of strict and uncertainty-expanded source prioritization as
    the number of cases that can be investigated increases from
    $k=2$ to $k=10$.}
    \label{fig:fig_7_decision_relevant_prioritization_instability}
\end{figure}

\section{Discussion}
\label{sec:discussion}

\subsection{Main Interpretation of the Findings}
\label{sec:discussion_integrated}

Two findings are central. First, a temporal prior learned from multi-disease
source data transferred to a real ANDV outbreak without target-specific
refitting. Second, transmission labels in real outbreak data were uncertain
enough to change inferred source structure and the identity of cases selected
for priority investigation. The ANDV benchmark provides the main external validation. The locked prior
achieved MRR 0.571 compared with 0.274 for the strongest fair source-trained
baseline, while median true-parent rank improved from 7 to 2
(Section~\ref{sec:results_andv_performance}). The result was stable to
candidate-window changes, bootstrap resampling, leave-one-task-out analysis,
and task-reversal diagnostics
(Section~\ref{sec:results_robustness}). Cross-disease evaluation on D1
supports the transfer interpretation: disease-macro MRR was 0.575 for the
learned prior and 0.452 for the Gaussian baseline, while the learned temporal
curve remained highly similar across held-out disease folds. The gain therefore
does not appear to depend on a single target-specific fit or on greater model
complexity. The uncertainty analyses address a different problem. In the NYC MPXV audit,
41 of 75 epidemiologically linked inter-host pairs (54.67\%) were unresolved
or not genomically supported as direct transmission. In ANDV and Guangdong
Delta, retaining plausible uncertain edges changed source rankings and top-5
priority sets. Top-5 Jaccard similarity between strict and expanded graphs was
0.667 for ANDV and 0.429 for Delta
(Table~\ref{tab:uncertainty_source_summary}). These results do not imply that
the expanded graph is the true graph, or that an uncertainty-aware ranking
would necessarily improve outbreak control. They show that treating uncertain
links as fixed ground truth can change which cases appear most important. The two findings are therefore connected but distinct. The temporal-prior
experiment concerns transfer of a ranking signal when only timing information
is available. The uncertainty analysis concerns the reliability of the labels
against which reconstruction methods are trained, evaluated, and interpreted.
Both matter because accurate inference is difficult to assess when the
benchmark itself contains unresolved transmission evidence.

\subsection{Relationship to Contemporary Reconstruction Methods}
\label{sec:discussion_regime}

The timing-only framework addresses a different operating regime from
genome-integrated transmission reconstruction. Methods such as
\texttt{outbreaker2}~\citep{campbell2018outbreaker2},
SCOTTI~\citep{de2016scotti}, and epi-genomic approaches
\citep{didelot2017genomic,carson2025mbe} can infer richer transmission
structures when sufficiently informative sequence data are available. They
incorporate evolutionary information, incomplete sampling, within-host
variation, and uncertainty that are outside the scope of a temporal ranking
model. The learned prior is intended for settings in which genomic evidence is absent,
delayed, sparse, or insufficient to resolve individual transmission pairs.
Its advantage is that it can be trained externally and applied immediately
without estimating an outbreak-specific interval distribution. Direct numerical
comparison with genome-based models on the ANDV timing-only benchmark would
therefore be inappropriate because the required inputs differ. A useful practical view is not to treat these approaches as competitors.
Timing-based ranking can provide an early source signal; genomic evidence can
refine transmission inference when available; and uncertainty in the resulting
links should remain visible rather than being converted automatically into
fixed labels. The MPXV and Delta analyses illustrate why this final step
matters even in data-rich outbreaks.

\subsection{Limitations}
\label{sec:5_3_discussion_limitations}

The ANDV benchmark uses high-confidence published transmission links as the
reference standard, but these links cannot be regarded as error-free direct
observations. The benchmark was deliberately restricted to cases with verified
onset dates, a unique high-confidence parent, and no documented competing
parent. The 2014 Argentina cluster also had genomic support for transmission
direction. These criteria make the ANDV labels stronger than the broader
epidemiological links examined in the MPXV audit, but residual label error
remains possible. Its effect on measured performance cannot be determined from
the available data and may depend on whether label error is related to temporal
gap. Accordingly, the reported $\mathrm{MRR}=0.571$ describes performance on the
best-characterized subset of ANDV transmission events; performance may be
lower for ambiguous multi-parent cases, incomplete temporal records, or
lower-confidence transmission links.

The second limitation is sample size. The primary external benchmark contains
29 ranked ANDV tasks, including two singleton candidate sets. Bootstrap,
jackknife, leave-one-task-out, candidate-window, and reversal analyses show that
the reported advantage is not driven by one or two influential tasks, but these
diagnostics do not replace validation in a larger independent outbreak. The
Global.Health audit also showed why such benchmarks are difficult to obtain:
among 134,095 screened records, no strict parent-ranking task with verified
parent and child onset dates could be constructed. Additional real-outbreak
benchmarks remain necessary. Performance also depends to some extent on the composition of the source
training data. Targeted source-domain ablations showed that ANDV performance changed when
MERS or Smallpox source data were removed, indicating some dependence on
source-domain composition without attributing the transfer result to a
single source analogue. The main conclusion is therefore
not that all source diseases contribute equally, but that the transfer result
was not explained by one obvious source-domain analogue or by randomized labels
alone.

The SVD analysis is exploratory and should not be interpreted as a second
external validation of the locked prior. Absolute onset dates were not uniformly
available, and the analysis used relative-time heuristics rather than the
serialized temporal-prior artifact. It therefore provides only evidence that
temporal proximity can be informative in that reconstructed network. The MPXV and Guangdong Delta modules also have limits of generalizability.
The MPXV analysis represents one urban outbreak with substantial contact
uncertainty and limited phylogenetic resolution. The Delta analysis represents
a separate pathogen and transmission setting, but still does not constitute a
representative sample of outbreak types. The evidence therefore supports the
existence of meaningful transmission-label uncertainty, not a universal estimate
of its prevalence across pathogens or surveillance systems.

Finally, the uncertainty framework compares discrete strict and expanded graph
configurations rather than estimating a posterior distribution over possible
transmission networks. Likewise, the fixed-capacity prioritization analysis is
a sensitivity analysis, not a prospective intervention study. Future work could
assign probabilistic weights or latent states to uncertain edges and propagate
that uncertainty directly through reconstruction and prioritization. Such an
approach would connect benchmark-level label uncertainty more closely with
probabilistic inference methods that already model uncertainty within the
reconstruction process.

\section{Conclusion}
\label{sec:conclusion}

A temporal prior learned from multi-disease source data transferred to a real
ANDV parent-ranking benchmark without target-specific refitting, improving MRR
from 0.274 to 0.571 relative to the strongest fair source-trained temporal
baseline. The advantage remained stable across cross-disease, finite-sample,
and candidate-window analyses. Independent outbreak evidence also showed
substantial uncertainty in published transmission links: more than half of the
audited MPXV inter-host links were genomically unresolved or unsupported, and
retaining uncertain links changed top-ranked source cases in both ANDV and
Guangdong Delta. These findings support two practical principles: temporal
information can provide a transferable early ranking signal, and uncertainty
in transmission labels should remain visible when reconstruction methods are
evaluated and used for source prioritization.

\section{Ethical Considerations and Data Sharing}
\label{sec:ethics}

All analyses used previously published outbreak data; no new participants were
recruited and no primary data were collected. ANDV case information came from
anonymized published materials. MPXV data came from the supplementary files of
Akther et al.~\citep{Akther2025NYCMpox}, whose study reports the relevant
ethical approvals and consent procedures. The Global.Health analysis used
openly accessible records, and the SVD pilot used published surveillance data.
No directly identifying information was retained.

\section{Code Availability}
\label{sec:code_availability}

A reference implementation is available as the open-source Python package
\texttt{temprior}, installable via \texttt{pip install temprior}. Source code
and documentation are available at
\url{https://github.com/01ahsan/TemPrior}. The public package provides a
general-use temporal prior consistent with the method described here; it is
not the exact locked model artifact used to produce the quantitative results
reported in this study.

\bibliographystyle{plainnat}
\bibliography{references}

\end{document}